\documentclass[11pt, journal, onecolumn]{IEEEtran}
\usepackage[letterpaper, left=1in, right=1in, bottom=1in, top=1in]{geometry}

\usepackage{xcolor,soul,framed} %,caption

\colorlet{shadecolor}{yellow}
\usepackage[pdftex]{graphicx}
\graphicspath{{../pdf/}{../jpeg/}}
\DeclareGraphicsExtensions{.pdf,.jpeg,.png}

\usepackage[cmex10]{amsmath}
\usepackage{array}
\usepackage{mdwmath}
\usepackage{mdwtab}
\usepackage{eqparbox}
\usepackage{url}
\usepackage{listings}
\usepackage{xcolor}
\usepackage{booktabs}
\usepackage{tabularx}
\usepackage{array}
\usepackage[flushleft]{threeparttable}
\usepackage{threeparttablex}
\usepackage{hyperref} % remove boxes with [hidelinks] preamble
\usepackage{lipsum}
\usepackage[ruled,vlined,linesnumbered]{algorithm2e}
\usepackage{footnote}
\usepackage{caption}
\usepackage{subcaption}
\usepackage{amsmath}
\usepackage{amssymb}% for \mathbb
\usepackage{stmaryrd}

\usepackage{lineno}
\usepackage{easylist}
\usepackage{amssymb}
\usepackage{amsmath}
\usepackage{subcaption}
\usepackage{url}
\usepackage{textcomp}
\usepackage{verbatim}
\usepackage[ruled,vlined]{algorithm2e}
\usepackage{ulem}
\usepackage{graphicx}
\usepackage{pgfplots}
\usepackage{listings} 
\pgfplotsset{compat=1.14}
\pgfplotsset{compat=newest}
\pgfplotsset{plot coordinates/math parser=false}
\usepackage{tikzscale}
\usetikzlibrary{matrix,chains,positioning,decorations.pathreplacing,arrows}
\usepackage{tikz-qtree,tikz-qtree-compat}
\usetikzlibrary{calc}
\modulolinenumbers[5]

\usepackage{caption}

\newcommand{\data}[1][n]{\mathcal{D}_{#1}}
\newcommand{\node}{t}

\lstdefinestyle{pythonstyle}{
    commentstyle=\color{codegreen},
    keywordstyle=\color{magenta},
    numberstyle=\tiny\color{codegray},
    stringstyle=\color{codepurple},
    basicstyle=\ttfamily\footnotesize,
    breakatwhitespace=false,         
    breaklines=true,                 
    captionpos=b,                    
    keepspaces=true,                 
    numbers=left,                    
    numbersep=5pt,                  
    showspaces=false,                
    showstringspaces=false,
    showtabs=false,                  
    tabsize=2
}

\begin{document}

\renewcommand\IEEEkeywordsname{Keywords} 

\bstctlcite{IEEEexample:BSTcontrol}
   % \title{\textit{Parea}: multi-view hierarchical ensemble clustering for disease subtype discovery}

% TITLE %%%%%%%%%%%%%%%%%%%%%%%
   
%\title{Smoothing tree ensembles for improved accuracy and interpretability}   
\title{Bayesian post-hoc regularization of \\ random forests}   
%\title{Blurred trees for improved classification accuracy of random forests}      

%%%%%%%%%%%%%%%%%%%%%%%

\DeclareRobustCommand*{\IEEEauthorrefmark}[1]{%
  \raisebox{0pt}[0pt][0pt]{\textsuperscript{\footnotesize\ensuremath{#1}}}}  

\author{\IEEEauthorblockN{Bastian Pfeifer}
%\IEEEauthorrefmark{1}$^{*}$}
%, Arne Gevaert\IEEEauthorrefmark{2} 
%and
%Markus Loecher\IEEEauthorrefmark{2} }

\IEEEauthorblockA{%\IEEEauthorrefmark{1}
Institute for Medical Informatics, Statistics and Documentation\\ Medical University Graz, Austria} \\
%\IEEEauthorrefmark{2}University of Ghent\\
%\IEEEauthorrefmark{2}Berlin School of Economy and Law
%}

\thanks{ 
%$^{*}$Corresponding author: Bastian Pfeifer (bastian.pfeifer@medunigraz.at).
}}

%\markboth{IEEE BIBM 2021 (\lowercase https://ieeebibm.org/BIBM2021/)}{}

\markboth{}{}

% ====================================================================
\maketitle

% === ABSTRACT ====================================================================
% =================================================================================
\begin{abstract}
%\boldmath
Random Forests are powerful ensemble learning algorithms widely used in various machine learning tasks. However, they have a tendency to overfit noisy or irrelevant features, which can result in decreased generalization performance. Post-hoc regularization techniques aim to mitigate this issue by modifying the structure of the learned ensemble after its training.

\noindent Here, we propose \textit{Bayesian post-hoc regularization} to leverage the reliable patterns captured by leaf nodes closer to the root, while potentially reducing the impact of more specific and potentially noisy leaf nodes deeper in the tree. This approach allows for a form of pruning that does not alter the general structure of the trees but rather adjusts the influence of leaf nodes based on their proximity to the root node. We have evaluated the performance of our method on various machine learning data sets. Our approach demonstrates competitive performance with the state-of-the-art methods and, in certain cases, surpasses them in terms of predictive accuracy and generalization.

\end{abstract}


% === KEYWORDS ====================================================================
% =================================================================================

% \fi
\IEEEpeerreviewmaketitle

% ====================================================================
% ====================================================================
% ====================================================================

% Introduction
\section{Introduction}
\label{sec:introduction}
Post-regularization techniques for random forests refer to methods used to reduce overfitting and improve the generalization performance of random forest models. Random forests are powerful ensemble learning algorithms that combine multiple decision trees to make predictions. However, they can still suffer from overfitting, especially when the trees in the forest become highly complex and tailored to the training data. Post-regularization techniques aim to address this issue by modifying or refining the random forest model after the initial training phase. These techniques typically focus on adjusting the complexity of individual trees or applying ensemble-level modifications. Some commonly used post-regularization techniques for random forests include:

\textbf{Pruning:} Pruning involves removing unnecessary branches or nodes from individual trees to simplify their structure. This helps prevent overfitting and promotes better generalization by reducing the complexity of the trees \cite{pmlr-v206-liu23h}.

\textbf{Feature selection:} Random forests can sometimes include irrelevant or redundant features, which can degrade performance. Feature selection techniques aim to identify and remove such features from the model, allowing it to focus on the most informative ones and potentially reducing overfitting \cite{kursa2010feature}\cite{pfeifer2022robust}.

%\textbf{Bagging:} Bagging is a technique that involves training multiple random forest models on different subsets of the training data and then combining their predictions. This helps to reduce variance and improve model generalization by leveraging the diversity among the individual models.

%\textbf{Stacking:} Stacking, or meta-ensemble learning, involves training a secondary model, often a simple one like logistic regression or a linear regression, to learn from the predictions of multiple random forest models. The secondary model combines the predictions of the random forests to make the final prediction, potentially improving overall performance.

\textbf{Calibration:} Calibration techniques aim to refine the predicted probabilities of random forests to better align with the true class probabilities. This can be particularly useful in tasks where reliable probability estimates are important, such as in certain risk assessment or medical diagnosis scenarios \cite{agarwal2022hierarchical}.

These post-regularization techniques provide various approaches to combat overfitting and enhance the generalization ability of random forest models. By incorporating these techniques into the random forest workflow, practitioners can often achieve better performance and more reliable predictions on unseen data.

Here, we present a Bayesian post-hoc regularization method for the calibration of the Dection trees' leaf node probabilities. Our method is implemented within the Python package \textit{TreeSmoothing}, which seemingly interfaces with sklearn functionalities and thus can be employed on any trained tree-based classifier.

%%%%%%%%%%%%%%%%%%%%%%%%%%%%%%%%%%%%%%%%%%%%%%%%%%%%%%%%%%%%%%%%%%%%%%%%%%%%%%%%%%%%%%%%%%%%%%
\section{Related Work - Hierarchical Shrinkage}

Agarwal et al. (2022) \cite{agarwal2022hierarchical} proposed a post-hoc regularization technique known as \textit{Hierarchical Shrinkage} (HS). Unlike modifying the tree structure, HS focuses on shrinking tree predictions and adjusting sample weights during training. This additional regularization improves generalization performance and allows for smaller ensembles without sacrificing accuracy. HS also enhances post-hoc interpretations by reducing noise in feature importance measures, leading to more reliable and robust interpretations. The method replaces the average prediction of a leaf node with a weighted average of the mean responses of the leaf and its ancestors, controlled by a regularization parameter $\lambda$ as defined in Eq.~(\ref{eq:HSdef}).

% max. 5 items with max. 85 Characters per item incl. spaces !
%The approaches summarized in section % \ref{sec:Introduction}
%Recently, \cite{agarwal2022hierarchical} proposed a {\it post-hoc} regularization referred to as {\it Hierarchical Shrinkage} (HS), which does not modify the tree structure. Instead, it shrinks the predictions of the trees or the sample weights used during training. The authors demonstrate that this additional regularization can further improve generalization performance. By reducing the complexity of the model through shrinkage, one can  effectively prevent overfitting and allow for the use of smaller ensembles for many datasets without sacrificing accuracy. 
%Furthermore, they find that HS also enhances the quality of post-hoc interpretations. Specifically, by reducing the noise in the feature importance measures, HSregularization seems to lead to more reliable and robust interpretations. 
%HS replaces the average prediction or response of a leaf node with a weighted average of the mean responses of the leaf and its ancestors. These weights are determined based on the number of samples in each leaf and are controlled by a single regularization parameter $\lambda$, as summarized in Eq.~(\ref{eq:HSdef}).

The following is a brief summary of the ideas proposed in \cite{agarwal2022hierarchical}.
Assume that we are given a training set $\data = (X;y)$. Our goal is to learn a tree model $\hat{f}$ that accurately represents the regression function based on this training data.
Given a query point $\mathbf{x}$, let $\node_L \subset \node_{L-1} \subset \cdots \subset \node_0$ denote its leaf-to-root path, with $\node_L$ and $\node_0$ representing its leaf node and the root node respectively.
For any node $\node$, let $N(\node)$ denote the number of samples it contains, and $\hat{\mathbb{E}}_{\node}\{y\}$ the average response.
The tree model prediction can be written as the telescoping sum

\begin{equation}
\hat{f}(\mathbf{x}) =  \hat{\mathbb{E}}_{t_0} \{y\} + \sum_{l=1}^L{\left( \hat{\mathbb{E}}_{t_l}\{y\} - \hat{\mathbb{E}}_{t_{l-1}}\{y\} \right)}
\end{equation}
HS transforms $\hat f$ into a shrunk model $\hat{f}_{\lambda}$ via the formula:
\begin{equation}
\hat{f}_{\lambda}(\mathbf{x}) := \hat{\mathbb{E}}_{t_0} \{y\} + \sum_{l=1}^L{ \frac{ \hat{\mathbb{E}}_{t_l}\{y\} - \hat{\mathbb{E}}_{t_{l-1}}\{y\} }{1 + \lambda/N(t_{l-1})} } ,
\label{eq:HSdef}
\end{equation}
where $\lambda$ is a hyperparameter chosen by the user, for example by cross validation. HS maintains the tree structure, and only modifies the prediction over each leaf node.

\section{Proposed approach: Bayesian post-hoc regularization}
\subsection{Intuition of approach}
The core idea of our approach is inspired by the pruning concept, where the complexity of the trees is reduced by decreasing its depth. Here, we do not aim to manipulate the general structure of the trees, but we are adopting the leaf node probabilities such that more weight is given to nodes near my the root node (pruning by calibration). According to this idea we propose to update a conjugate Beta prior $\mathbf{B}_{prior}(\alpha, \beta)$ from the root node to the leaf nodes by subsequently adding the number of classified samples to the model parameters $\alpha$ (class 0) and $\beta$ (class 1). The leaf node probabilities are determined using the probabilities to observing a specific class given the inferred posterior Beta distribution $\mathbf{B}_{posterior}(\alpha, \beta)$.  

\subsection{Mathematical formulation}

\begin{equation}
    {\mathbf{B}}_{posterior}(\alpha, \beta) = {\mathbf{B}}_{prior}(\alpha, \beta) +  
     \sum_{l=0}^L{\mathbf{B}_{t_{l}}(\alpha + N_{0}(t_{l}) , \beta + N_{1}(t_{l}))},
\end{equation}
where $N_{0}(t_{l})$ refers to the number of samples classified as class 0, and $N_{1}(t_{l})$ is the number of samples classified as class 1 at node $t_{l}$.

The leaf node probabilities are calculate as 
\begin{equation}
\hat{f}_{\alpha,\beta}(\mathbf{x}) = \mathbf{PPF}(\frac{\alpha}{\alpha + \beta}| {\mathbf{B}}_{posterior}(\alpha, \beta)),
\end{equation}
where $\mathbf{PPF}$ is  percent point function (inverse of the cumulative distribution function - percentiles). 

% Results
\section{Evaluation}

\subsection{Data sets}
We assessed the accuracy of our proposed methodology on four machine learning benchmark datasets (see Table \ref{tab:table1}), downloaded using the Python packages imodels  \cite{Singh2021} and PMLB \cite{romano2021pmlb}. 

\begin{table}[h!]
  \begin{center}
    \caption{Benchmark datasets}
    \label{tab:table1}
    \begin{tabular}{l l l l l} % <-- Alignments: 1st column left, 2nd middle and 3rd right, with vertical lines in between
      \textbf{Datasets} & Sample size & Features & Class 0 & Class 1 \\
      \hline
      Breast cancer &  286 & 9 & 196 & 81 \\
      Habermann    & 306 & 3 & 81 & 225  \\
       Heart & 270 & 15 & 150 & 120 \\ 
      Diabetes & 768 & 8 & 500 & 268 \\
      \hline
    \end{tabular}
  \end{center}
  \centering
%Display of the median Cox log-rank $p$-values. Significant results ($\alpha=0.05$) are highlighted in bold. 
\end{table}

\subsection{Evaluation strategy}
In an first experiment, we used grid search-based on 5-fold crossvalidation to infer the optimal values for the hyperparameters. Following we performed 5-fold crossvalidation and calculated the mean accuracy. The aforementioned procedure was repeated 20 times and we report on results based on balanced accuracy and ROC-AUC.

In an second experiment, we have splitted the data into a train and test dataset. On the train dataset 5-fold crossvalidation was performed to tune the hyperparameters. The tuned model was then tested on the independent test data set. The described procedure was repeated 20 times and we again report results based on balanced accuracy and ROC-AUC as a performance metric.

In case of the herein \textit{Bayesian regularization} technique, the model-specific $\alpha$ and $\beta$ hyperparameters were grid-searched within $[1500, 1000, 800, 500, 100, 50, 30, 10, 1]$. For the \textit{Hierarchical Shrinkage} method we used $\lambda=[0.001, 0.01, 0.1, 1, 10, 25, 50, 100, 200]$. 

\subsection{Metrics used for evaluation}
The \textit{Balanced Accuracy} is a performance metric used in classification tasks to measure the accuracy of a model by taking into account the imbalance in the class distribution. It is defined as the average of the per-class accuracies, where the per-class accuracy is the ratio of the correctly classified instances to the total number of instances in that class. The formula for calculating Balanced Accuracy is given by:

\begin{equation}
\text{Balanced Accuracy} = \frac{1}{N_c} \sum_{i=1}^{N_c} \frac{TP_i}{TP_i + FN_i}
\end{equation}

where $N_c$ is the number of classes, $TP_i$ represents the number of true positive instances in class $i$, and $FN_i$ represents the number of false negative instances in class $i$. The Balanced Accuracy ranges from 0 to 1, where 1 indicates perfect classification performance and 0 indicates random classification.

The \textit{Receiver Operating Characteristic Area Under the Curve} (ROC-AUC) quantifies the performance of a model by measuring the area under the receiver operating characteristic curve. The ROC-AUC is computed as the area under the ROC curve, which ranges from 0 to 1. An ROC-AUC score of 1 indicates a perfect classifier, while a score of 0.5 suggests a random classifier. A higher ROC-AUC score indicates better discriminative ability of the model in distinguishing between the positive and negative classes.

To calculate the ROC-AUC, various methods can be used, including the trapezoidal rule or the Mann-Whitney U statistic. The formula for calculating ROC-AUC using the trapezoidal rule is given by:

\begin{equation}
\text{ROC-AUC} = \int_{0}^{1} \text{TPR}(FPR^{-1}(t)) \, dt
\end{equation}

where $\text{TPR}(FPR^{-1}(t))$ represents the true positive rate at the threshold corresponding to the inverse of the false positive rate $t$, and $FPR^{-1}(t)$ is the inverse of the false positive rate. The ROC-AUC provides a single scalar value that summarizes the model's performance across all possible classification thresholds. 

The difference between of the two metrics is that Balanced Accuracy focuses on the accuracy of individual classes, accounting for class imbalance, while ROC-AUC evaluates the overall discriminative ability of a classifier across all possible classification thresholds, providing a single scalar value. Both metrics have their own significance and are used in different contexts based on the requirements of the classification problem at hand.

\section{Results and discussion}
The results based on the herein analyzed four bechmark datasets suggests that our approach is competitive with hierarchical shrinkage in terms of the ROC-AUC values. When Balanced Accuracy is used as a performance metric, however, our Bayesian post regularization procedure is superior in most cases (see Figures 1-4). In all cases the post regularization techniques under investigation clearly can improve the accuracy of the vanilla random forest. Currently, we are working on an in-depth evaluation of the herein proposed Bayesian regularization techniques, while enhancing our method also to regression trees.   

% BREAST CANCER
\begin{figure}
     \centering
     \begin{subfigure}[b]{0.49\textwidth}
         \centering
         \includegraphics[width=\textwidth]{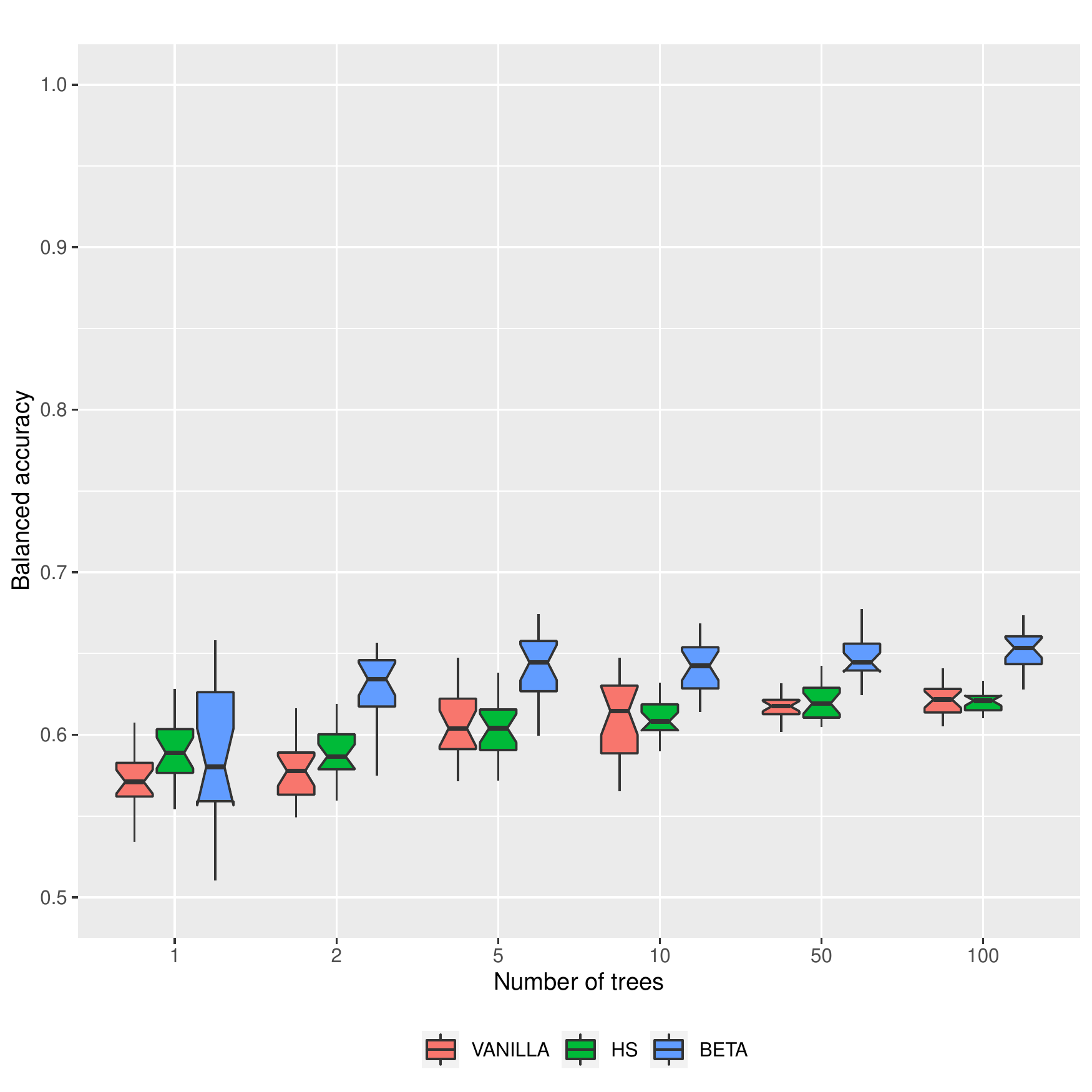}
         \caption{Balanced accuracy}
        % \label{fig:y equals x}
     \end{subfigure}
     %\hfill
     \begin{subfigure}[b]{0.49\textwidth}
         \centering
         \includegraphics[width=\textwidth]{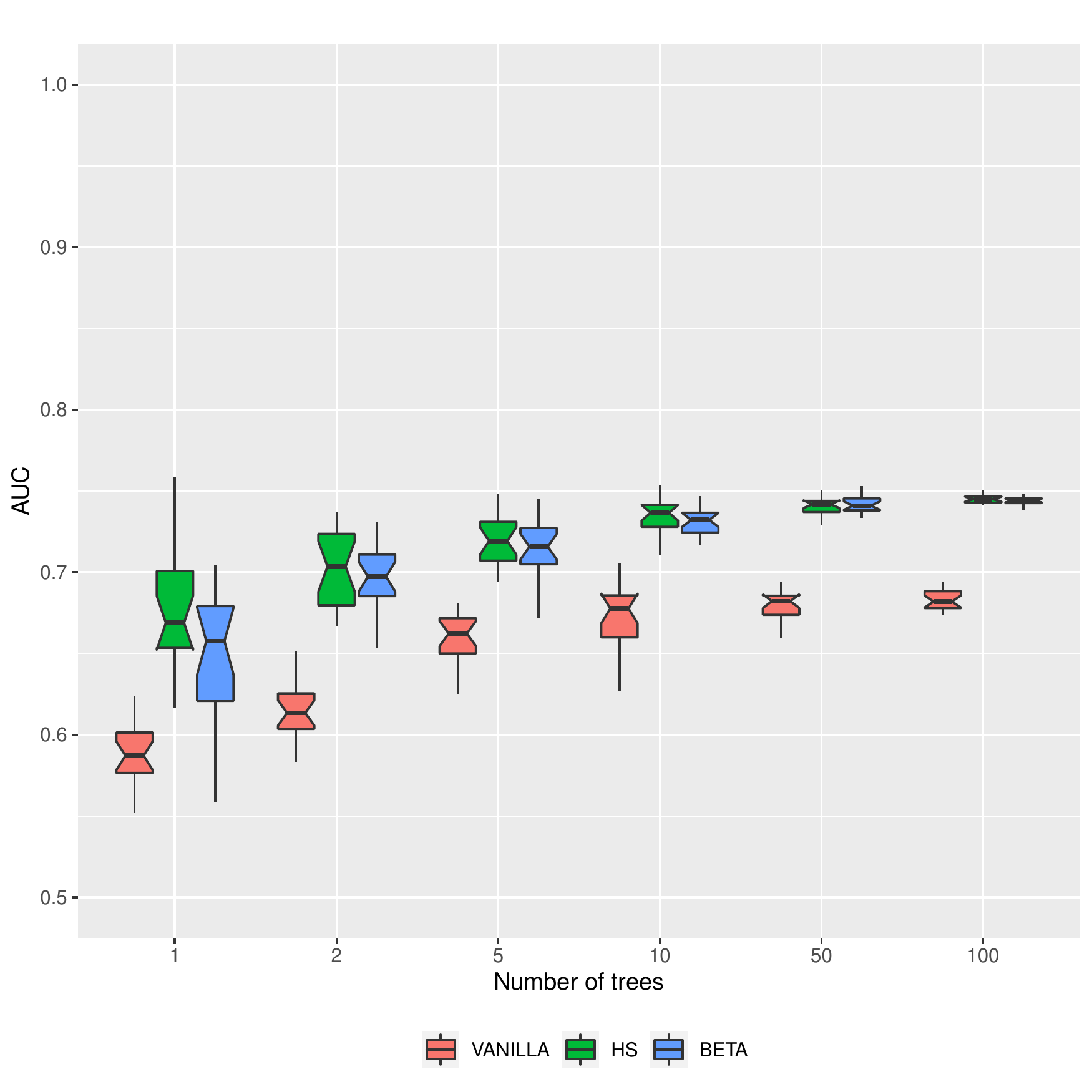}
         \caption{AUC}
         %\label{fig:three sin x}
     \end{subfigure}
       % \caption{Breast cancer dataset. 20 times 5-fold crossvaliation on the whole dataset.}
     %
     \begin{subfigure}[b]{0.49\textwidth}
         \centering
         \includegraphics[width=\textwidth]{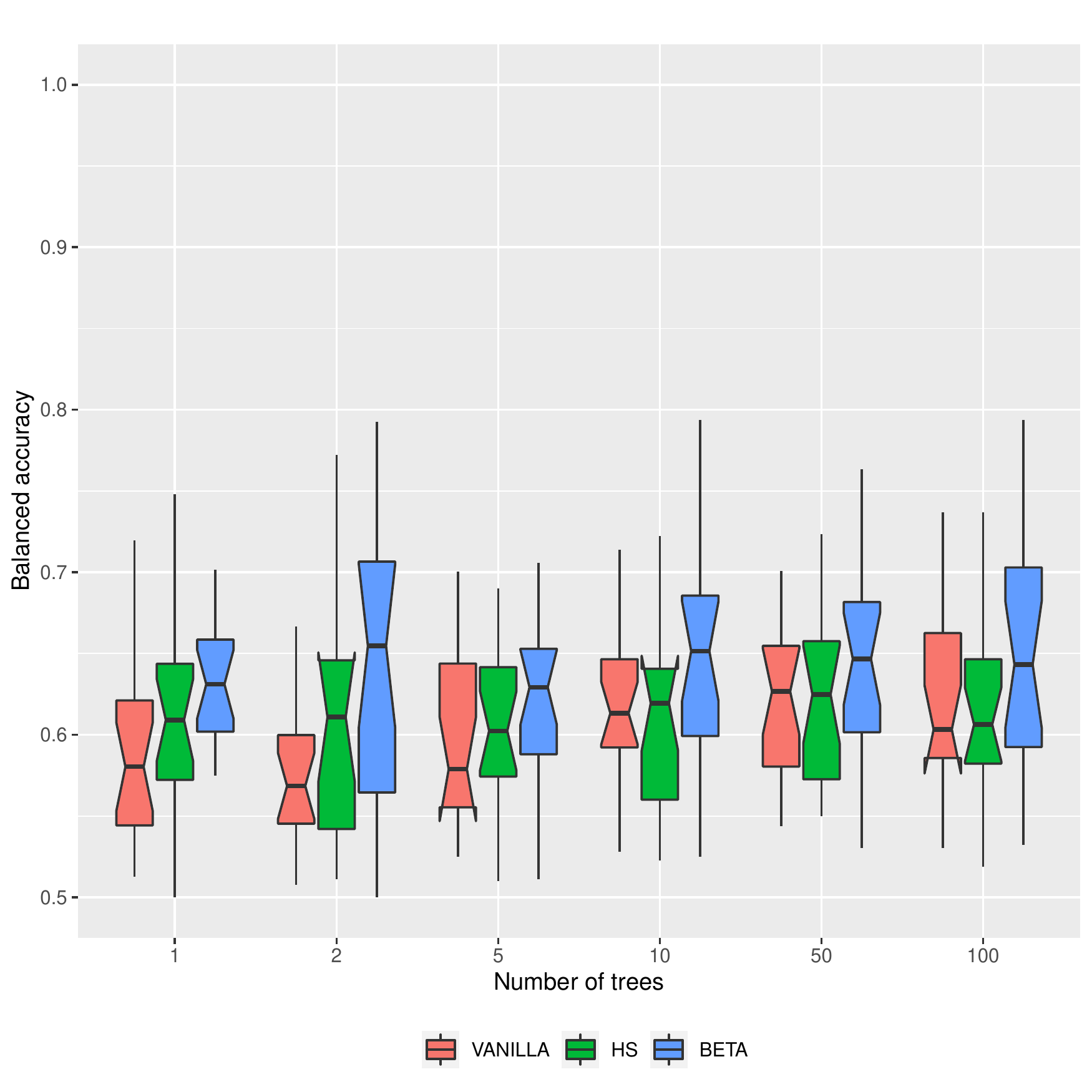}
         \caption{Balanced accuracy}
        % \label{fig:y equals x}
     \end{subfigure}
     %\hfill
     \begin{subfigure}[b]{0.49\textwidth}
         \centering
         \includegraphics[width=\textwidth]{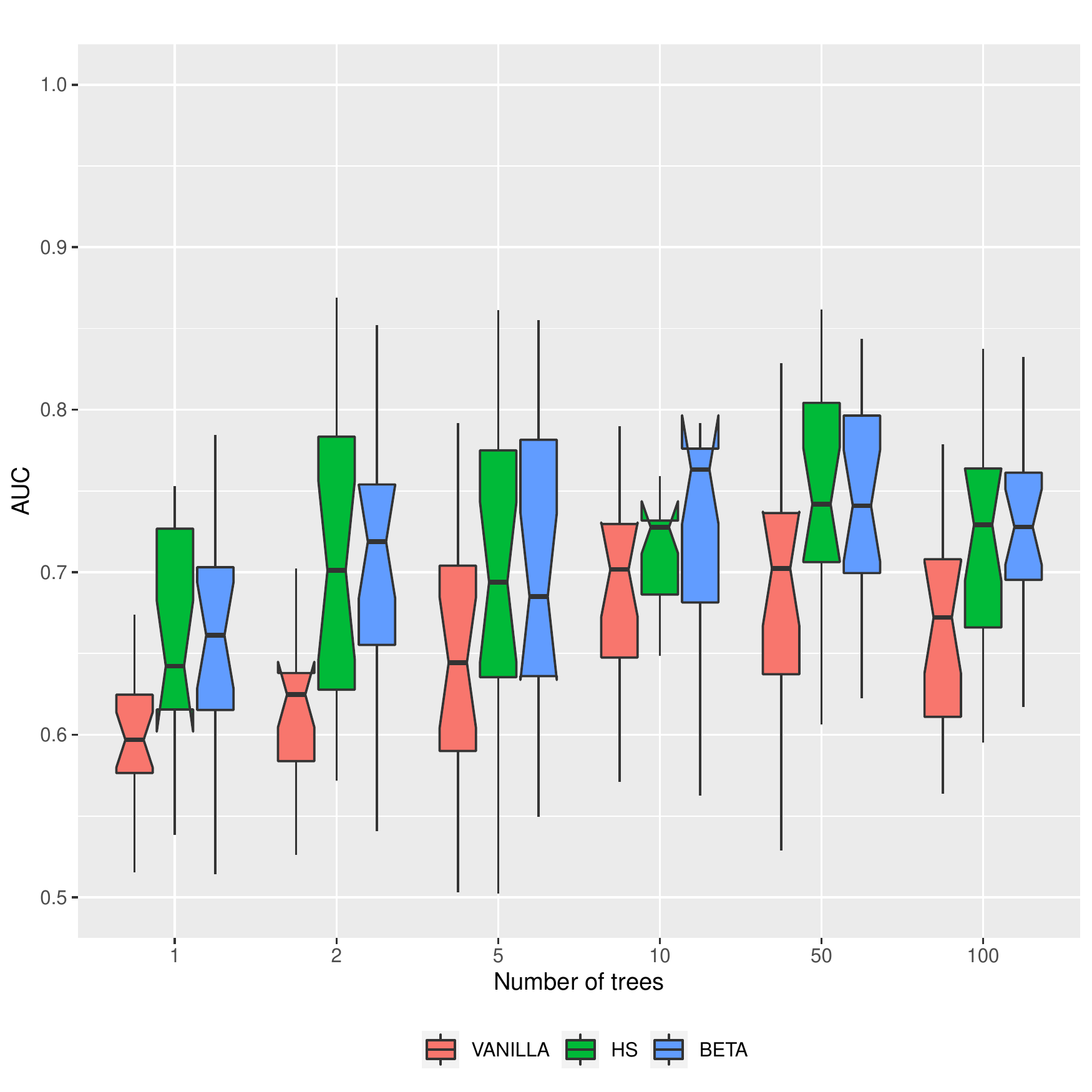}
         \caption{AUC}
         %\label{fig:three sin x}
     \end{subfigure}
        \caption{\textbf{Breast cancer dataset.} (a) and (b) 20 times 5-fold crossvaliation on the whole dataset. (c) and (d) 20 times evaluation on independent test dataset.}
        
        \label{fig:Breast}
\end{figure}

% HABERMANN
\begin{figure}
     \centering
     \begin{subfigure}[b]{0.49\textwidth}
         \centering
         \includegraphics[width=\textwidth]{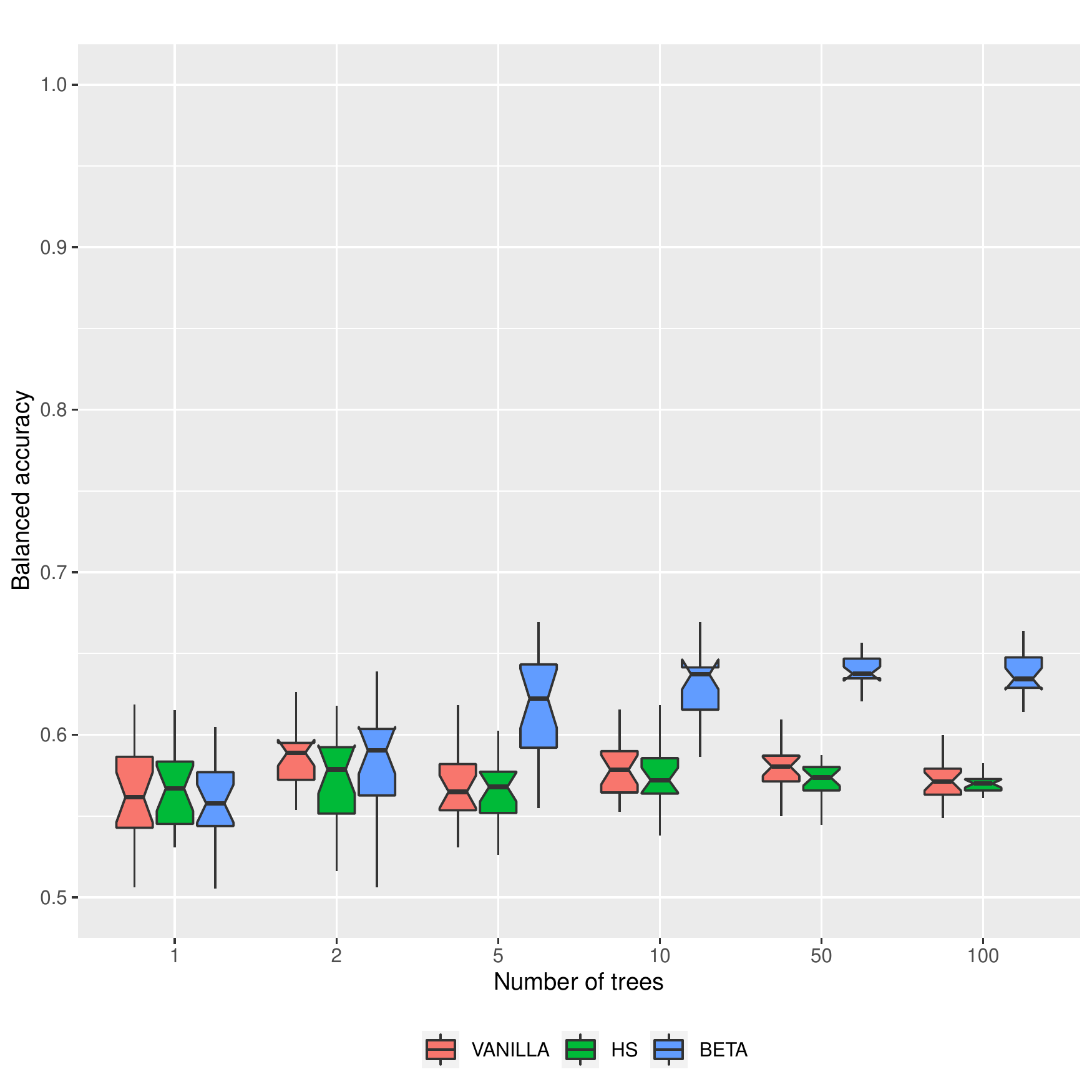}
         \caption{Balanced accuracy}
        % \label{fig:y equals x}
     \end{subfigure}
     %\hfill
     \begin{subfigure}[b]{0.49\textwidth}
         \centering
         \includegraphics[width=\textwidth]{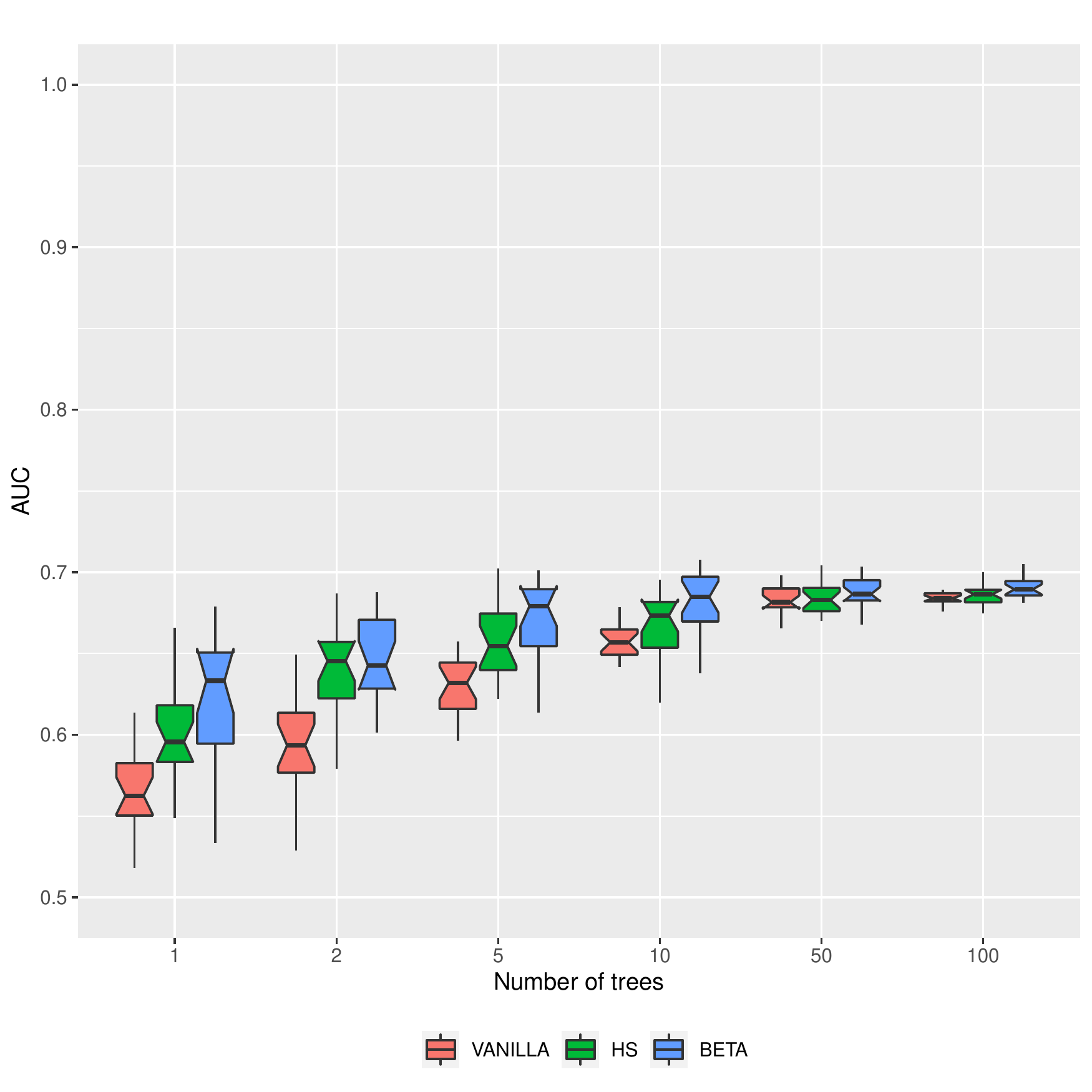}
         \caption{AUC}
         %\label{fig:three sin x}
     \end{subfigure}
     \begin{subfigure}[b]{0.49\textwidth}
         \centering
         \includegraphics[width=\textwidth]{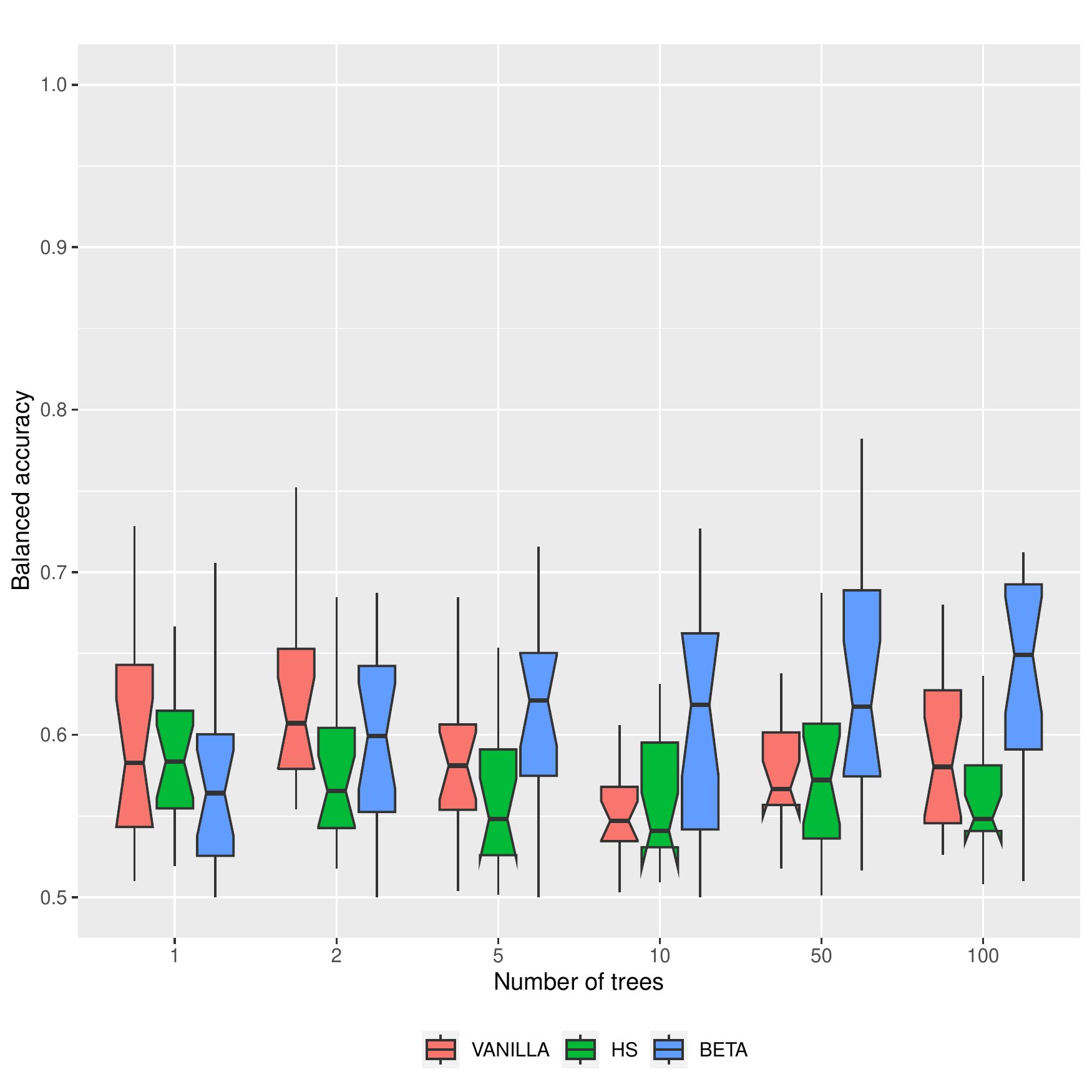}
         \caption{Balanced accuracy}
        % \label{fig:y equals x}
     \end{subfigure}
     %\hfill
     \begin{subfigure}[b]{0.49\textwidth}
         \centering
         \includegraphics[width=\textwidth]{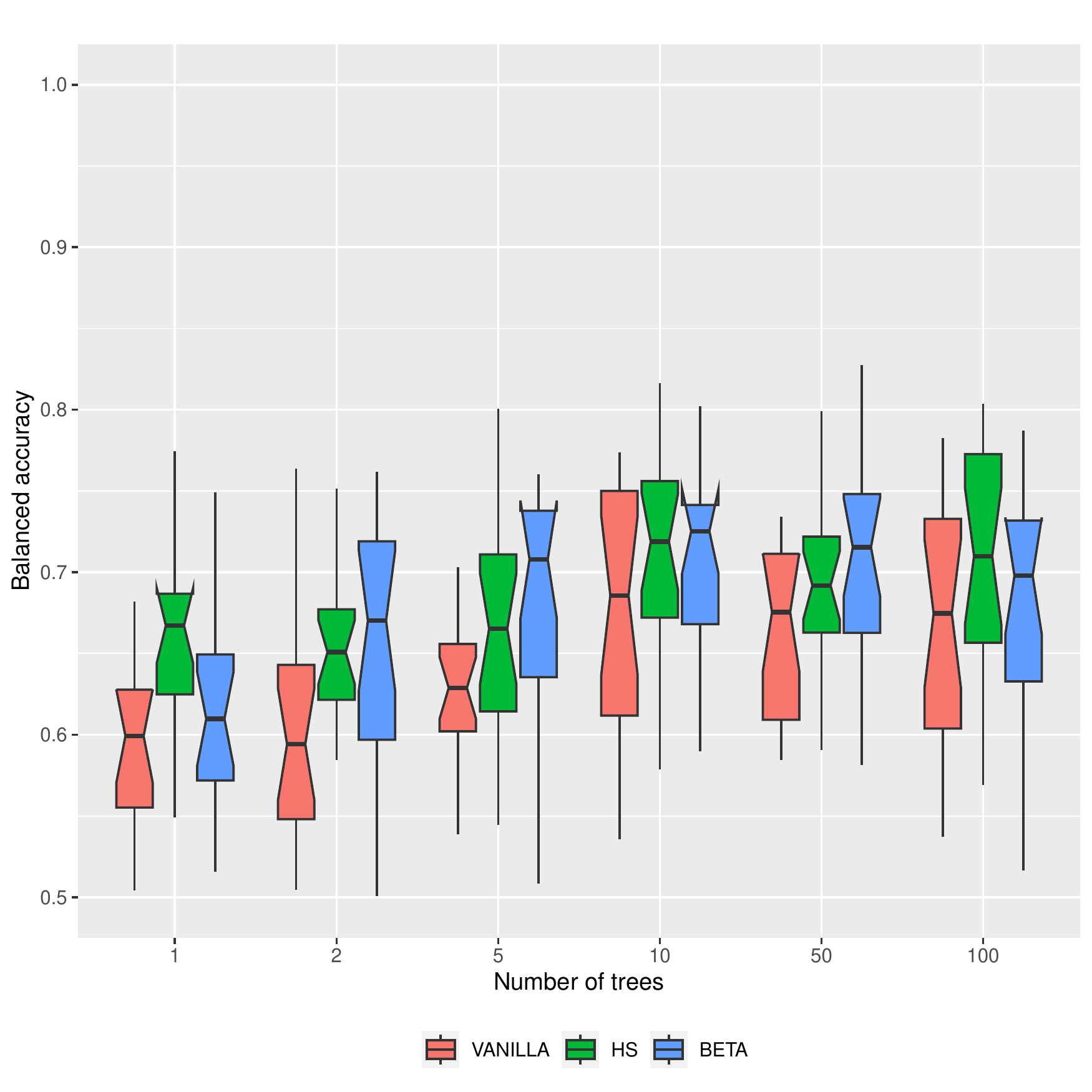}
         \caption{AUC}
         %\label{fig:three sin x}
    \end{subfigure}
        \caption{\textbf{Habermann dataset.} (a) and (b) 20 times 5-fold crossvaliation on the whole dataset. (c) and (d) 20 times evaluation on independent test dataset.}
        \label{fig:IRIS}
\end{figure}

% HEART CANCER
\begin{figure}
     \centering
     \begin{subfigure}[b]{0.49\textwidth}
         \centering
         \includegraphics[width=\textwidth]{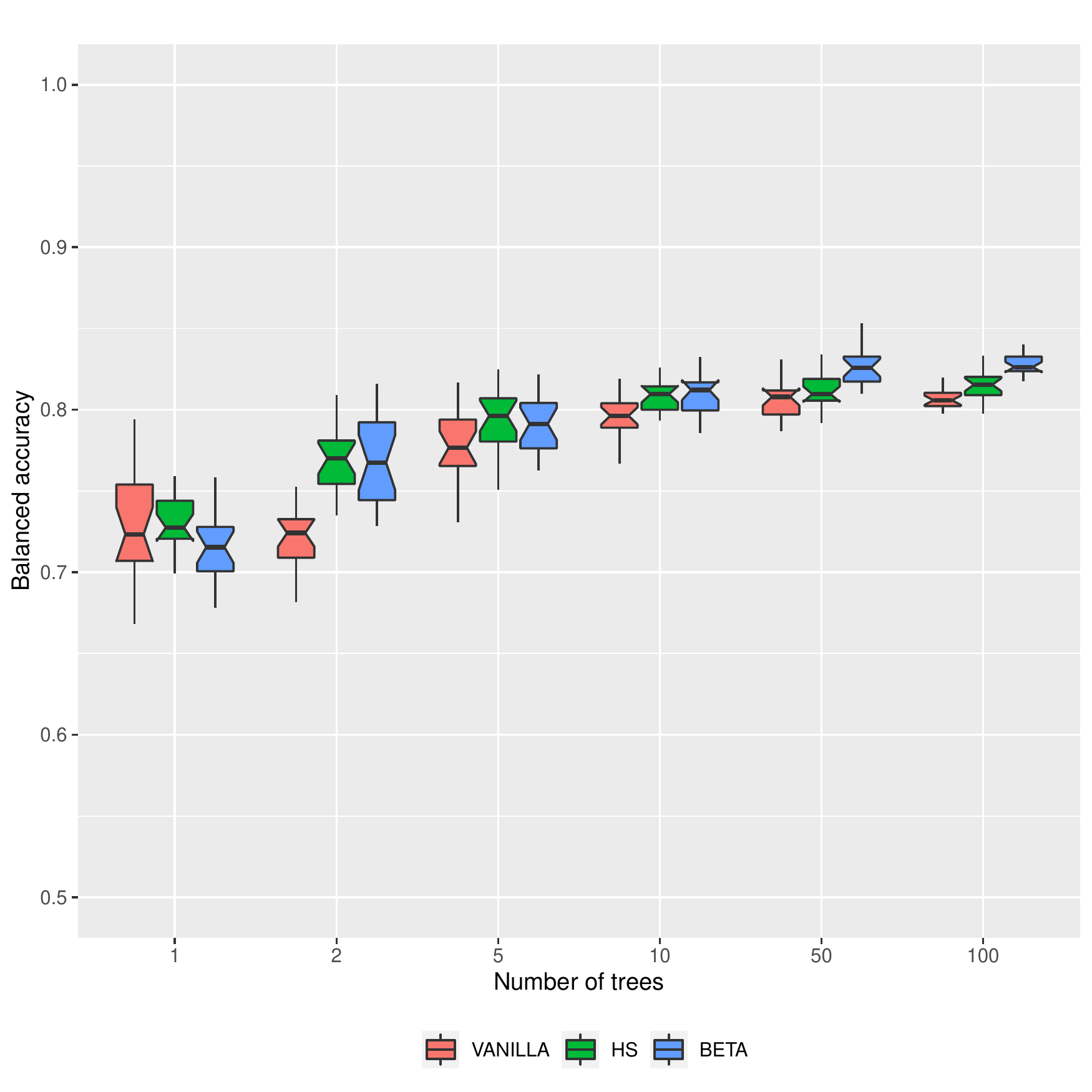}
         \caption{Balanced accuracy}
        % \label{fig:y equals x}
     \end{subfigure}
     %\hfill
     \begin{subfigure}[b]{0.49\textwidth}
         \centering
         \includegraphics[width=\textwidth]{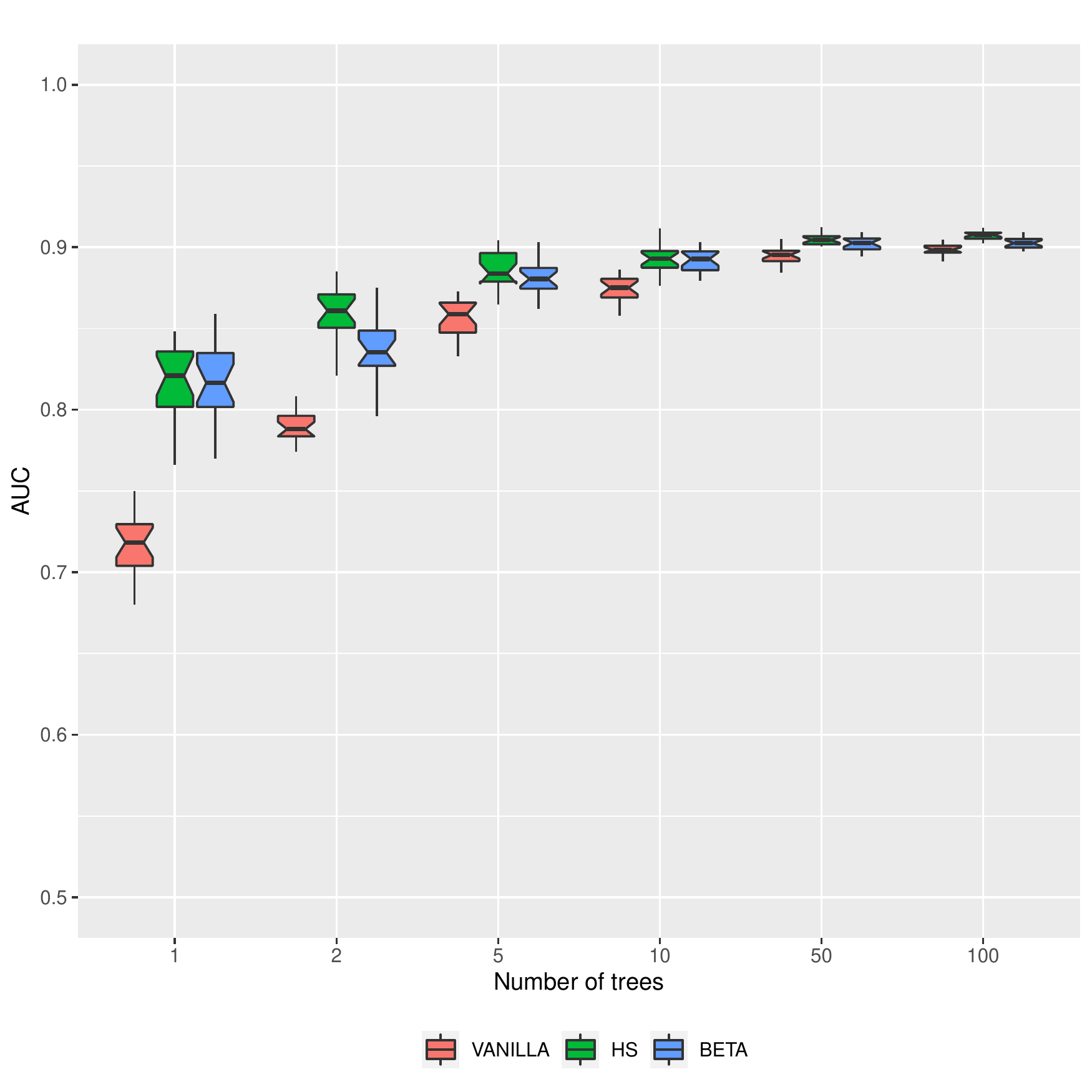}
         \caption{AUC}
         %\label{fig:three sin x}
     \end{subfigure}
    \begin{subfigure}[b]{0.49\textwidth}
         \centering
         \includegraphics[width=\textwidth]{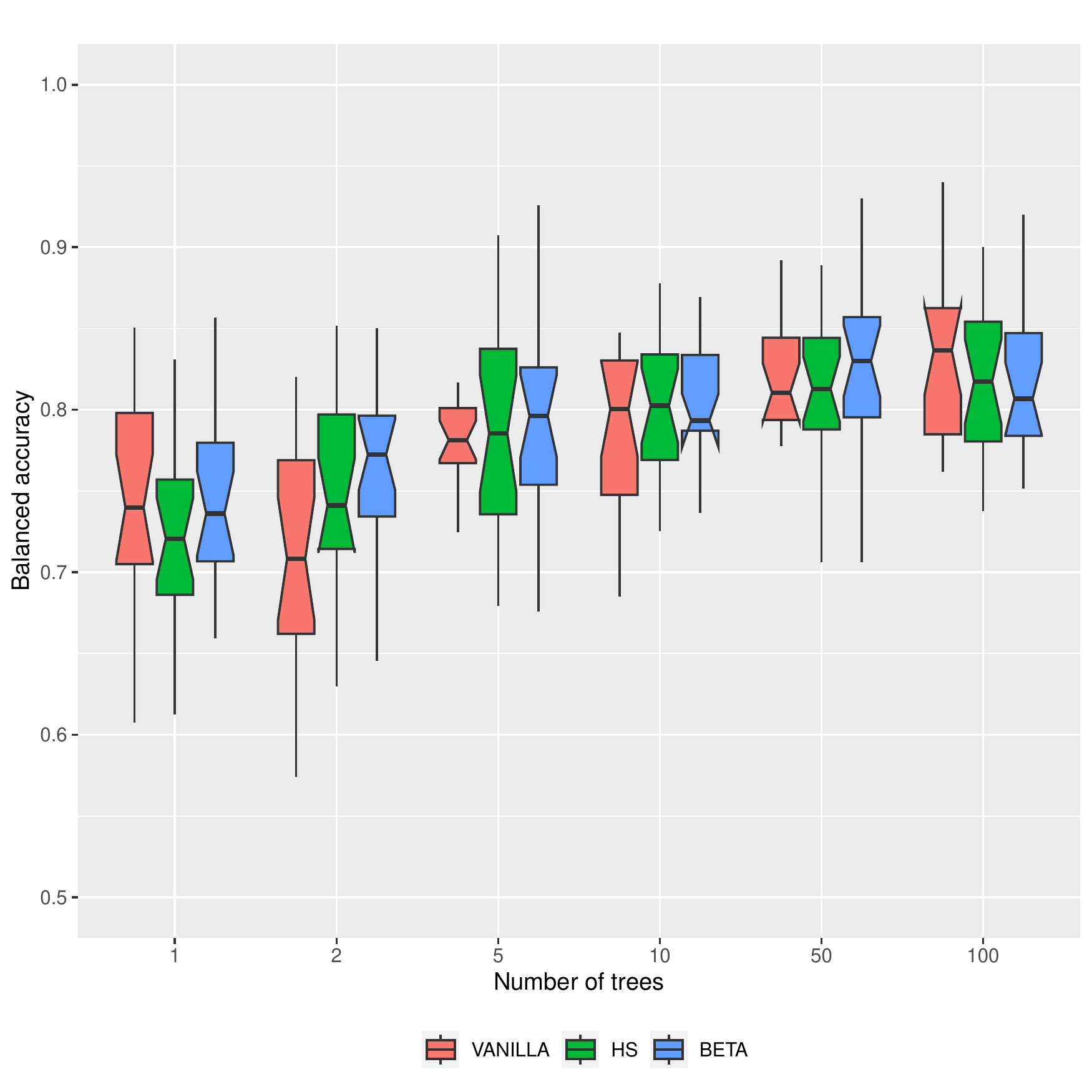}
         \caption{Balanced accuracy}
        % \label{fig:y equals x}
     \end{subfigure}
     %\hfill
     \begin{subfigure}[b]{0.49\textwidth}
         \centering
         \includegraphics[width=\textwidth]{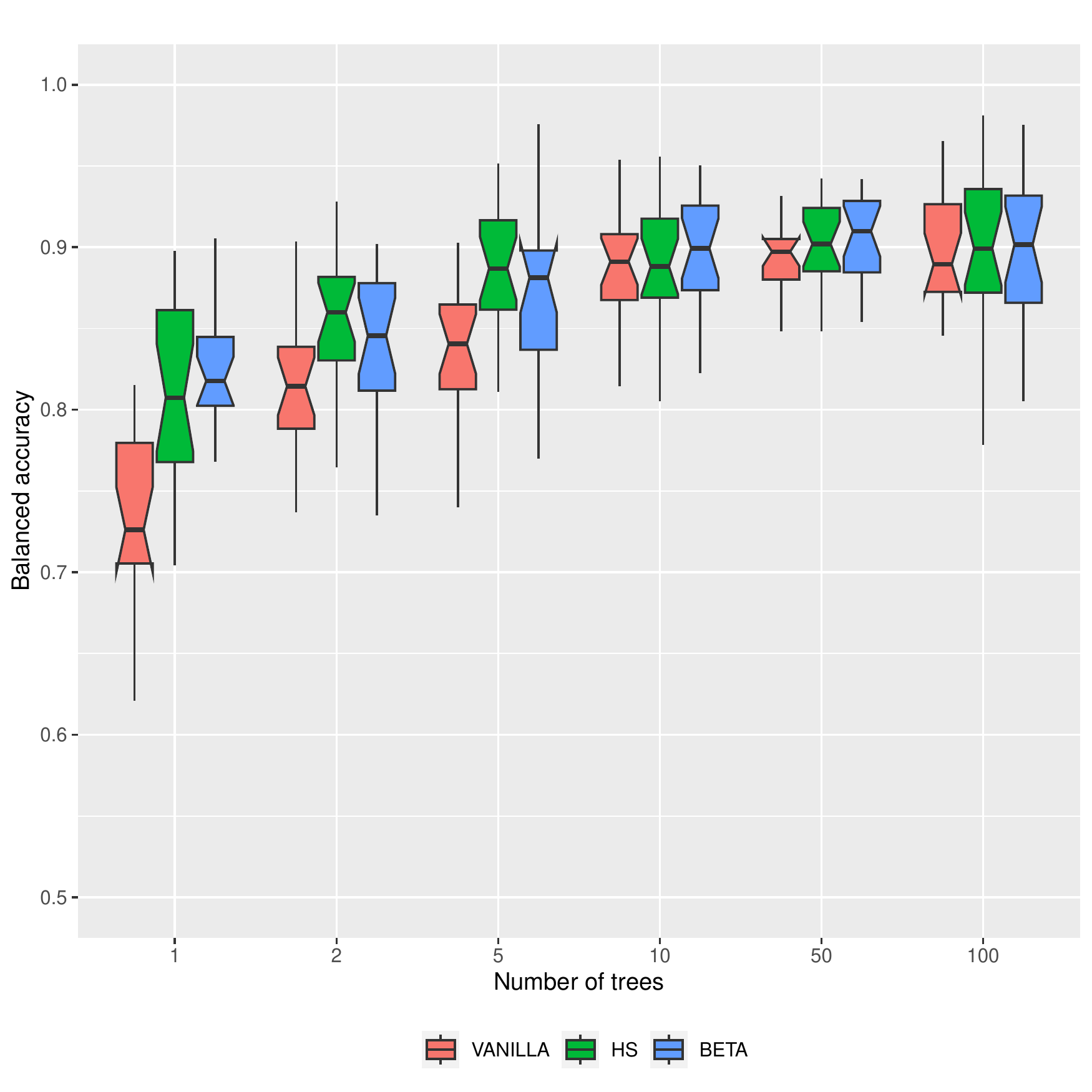}
         \caption{AUC}
         %\label{fig:three sin x}
     \end{subfigure}   
        \caption{\textbf{Heart dataset.} 20 times 5-fold crossvaliation on the whole dataset. (c) and (d) 20 times evaluation on independent test dataset.}
        \label{fig:Heart}
\end{figure}

% DIEABETES
\begin{figure}
     \centering
     \begin{subfigure}[b]{0.49\textwidth}
         \centering
         \includegraphics[width=\textwidth]{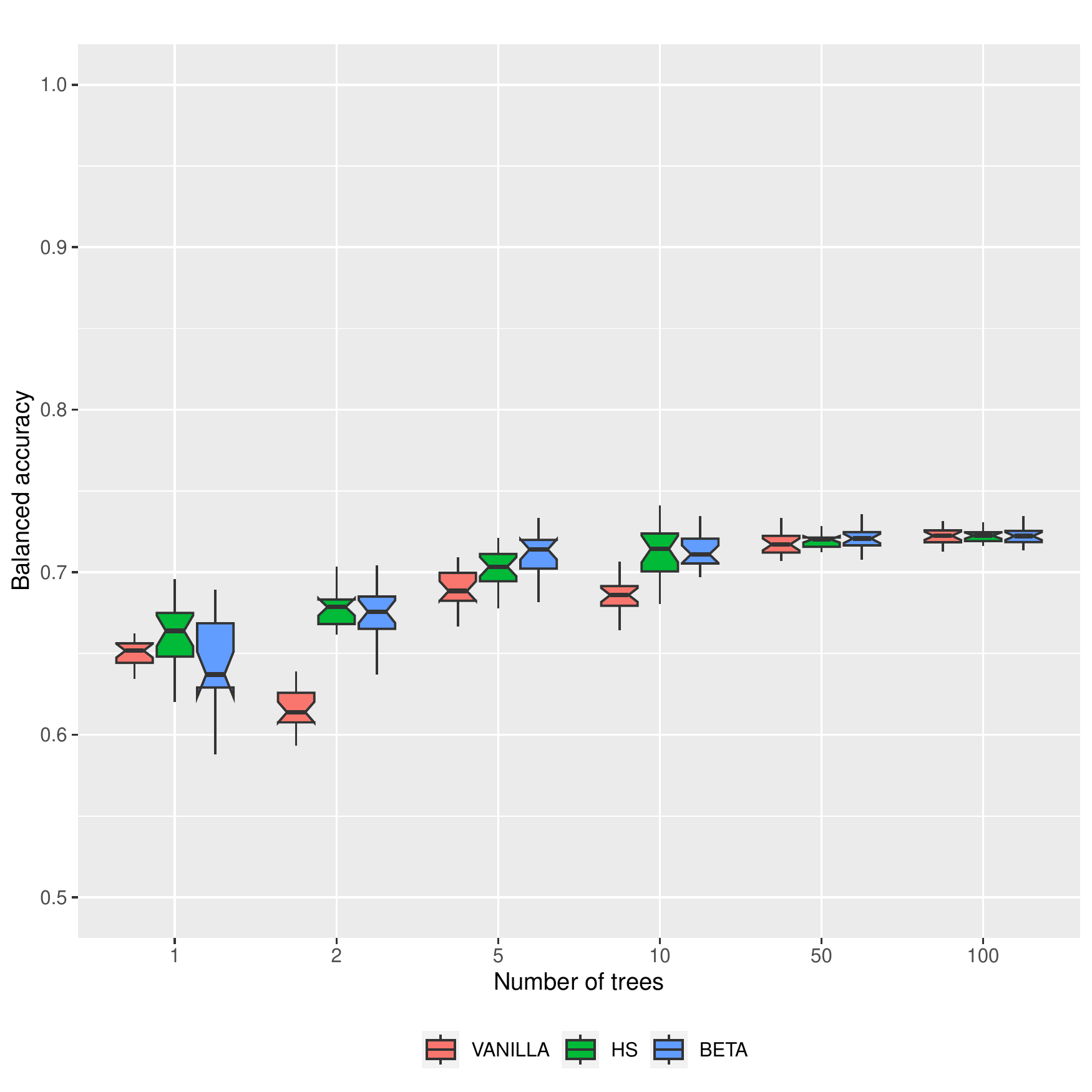}
         \caption{Balanced accuracy}
        % \label{fig:y equals x}
     \end{subfigure}
     %\hfill
     \begin{subfigure}[b]{0.49\textwidth}
         \centering
         \includegraphics[width=\textwidth]{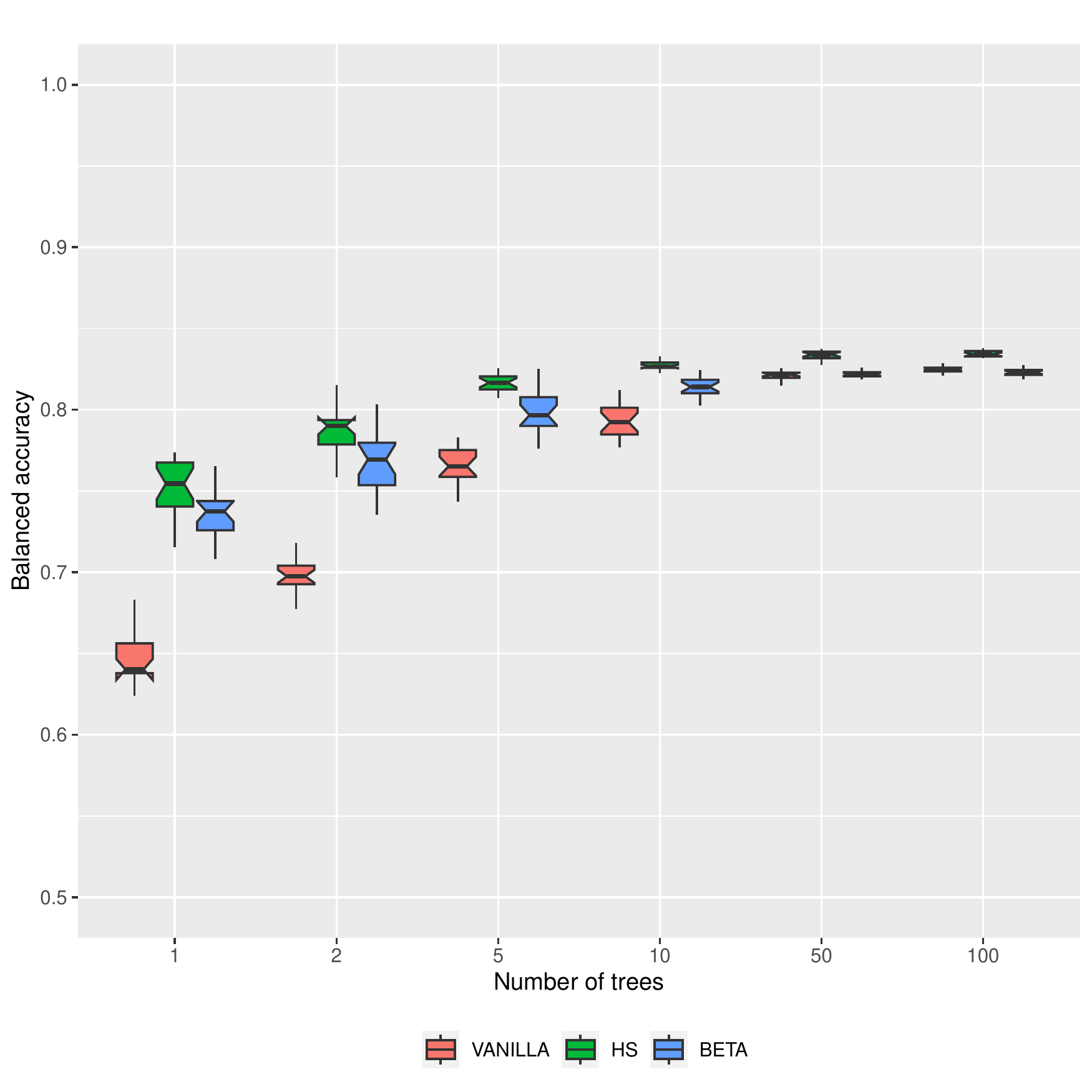}
         \caption{AUC}
         %\label{fig:three sin x}
     \end{subfigure}
    \begin{subfigure}[b]{0.49\textwidth}
         \centering
         \includegraphics[width=\textwidth]{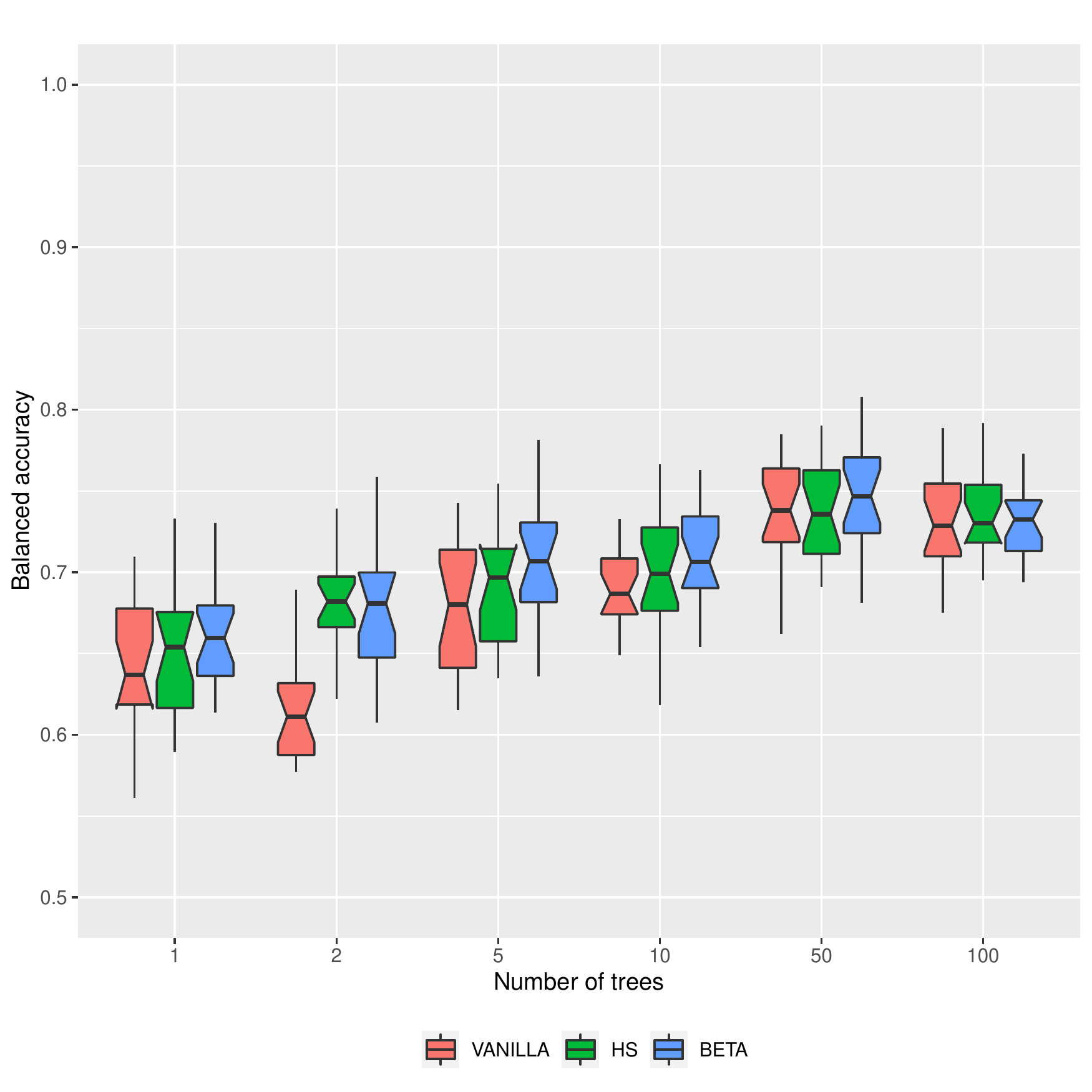}
         \caption{Balanced accuracy}
        % \label{fig:y equals x}
     \end{subfigure}
     %\hfill
     \begin{subfigure}[b]{0.49\textwidth}
         \centering
         \includegraphics[width=\textwidth]{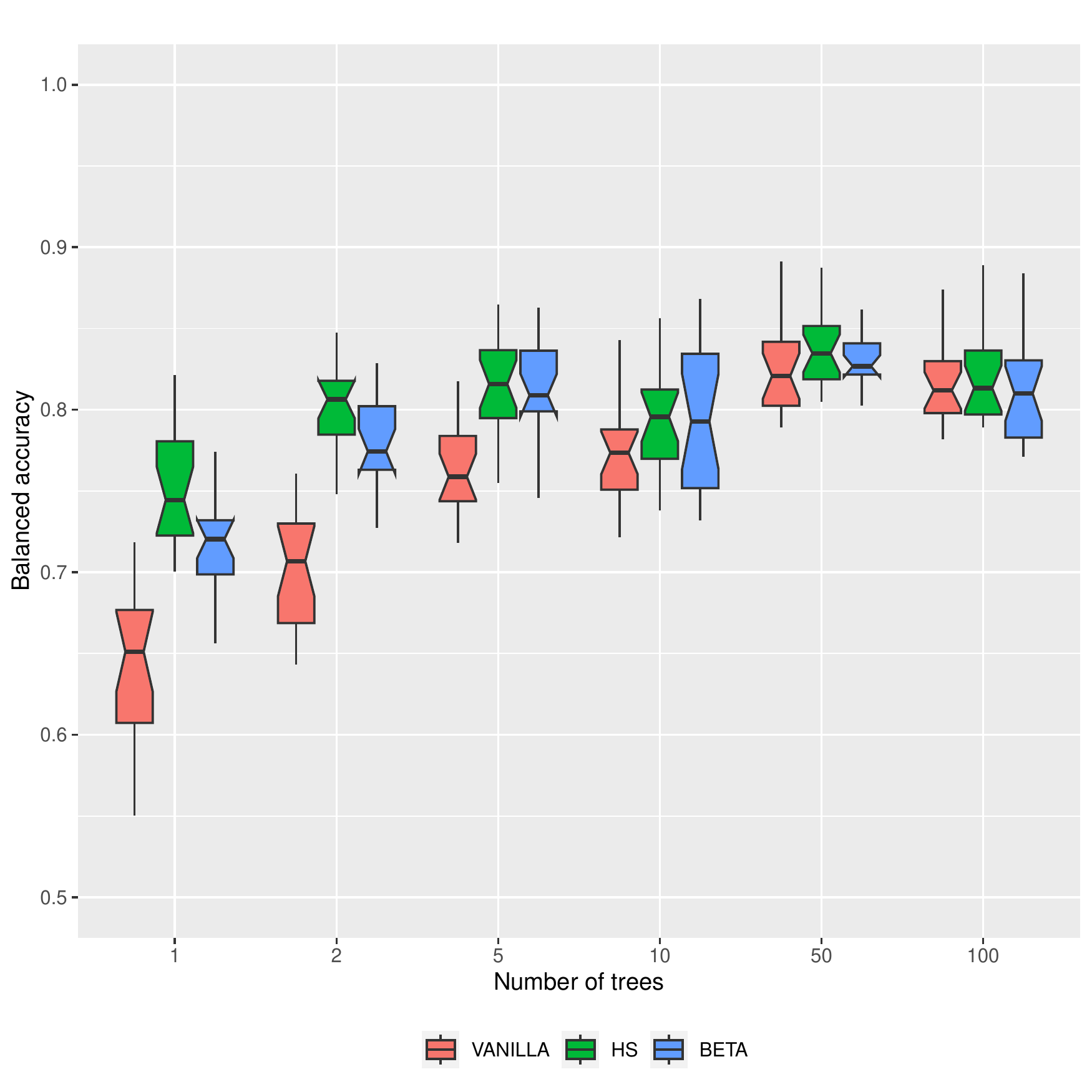}
         \caption{AUC}
         %\label{fig:three sin x}
     \end{subfigure}   
        \caption{\textbf{Diabetes dataset.} 20 times 5-fold crossvaliation on the whole dataset. (c) and (d) 20 times evaluation on independent test dataset.}
        \label{fig:Diabet}
\end{figure}

\label{sec:results}

\section{Data availability}
The proposed methodology is implemented within the Python package \textit{TreeSmoothing}, freely available on GitHub (\url{https://github.com/pievos101/TreeSmoothing}).

\section{Acknowledgments}
We would like to thank Arne Gevaert, Martin Uhrschler, and Markus Loecher for helpful discussions.

%\section*{Abbreviations}

%\begin{itemize}

%\item HC = Hierarchical Clustering
%\item UPGMA = Unweighted Pair-Group Method using Arithmetic Averages
%\item WPGMA =  Weighted Pair-Group Method using Arithmetic Averages
%\item WPGMC = Weighted Pair-Group Method using Centroids
%\item UPGMC = Unweighted Pair-Group Method using Centroids
%\item SIL = Silhouette Coefficient
%\item TCGA = The Cancer Genome Atlas
%\item GBM = Glioblastoma Multiforme
%\item KIRC = Kidney Renal Clear Cell Carcinoma
%\item LIHC = Liver Hepatocellular Carcinoma
%\item SARC = Sarcoma
%\item RNA =  Ribonucleic Acid

%\end{itemize}

%%%%%%%%%%%%%%%%%%%%%%%%%%%%%%%%%%%%%%

%\section*{Acknowledgments}
%\hl{We would like to thank ...}

\bibliographystyle{IEEEtran}
\bibliography{IEEEabrv,Bibliography}

\vfill


\end{document}